\documentclass[conference]{IEEEtran}
\IEEEoverridecommandlockouts
\usepackage{cite}
\usepackage{amsmath,amssymb,amsfonts}
\usepackage{algorithmic}
\usepackage{graphicx}
\usepackage{textcomp}
\usepackage{xcolor}
\usepackage{multirow}
\usepackage{import}
\usepackage{hyperref}
\usepackage{CJKutf8}
\usepackage{verbatim}
\def\BibTeX{{\rm B\kern-.05em{\sc i\kern-.025em b}\kern-.08em
    T\kern-.1667em\lower.7ex\hbox{E}\kern-.125emX}}
    
\begin{document}

\title{LoRA-like Calibration for Multimodal Deception Detection using ATSFace Data}



\author{\IEEEauthorblockN{Shun-Wen Hsiao}
\IEEEauthorblockA{\textit{Dept. of Management Information Systems} \\
\textit{National Chengchi University}\\
Taipei, Taiwan \\
hsiaom@nccu.edu.tw}
\and
\IEEEauthorblockN{Cheng-Yuan Sun}
\IEEEauthorblockA{\textit{Dept. of Management Information Systems} \\
\textit{National Chengchi University}\\
Taipei, Taiwan \\
110356041@nccu.edu.tw}
}

\maketitle

\begin{abstract}

Recently, deception detection on human videos is an eye-catching techniques and can serve lots applications. AI model in this domain demonstrates the high accuracy, but AI tends to be a non-interpretable black box. We introduce an attention-aware neural network addressing challenges inherent in video data and deception dynamics. This model, through its continuous assessment of visual, audio, and text features, pinpoints deceptive cues. We employ a multimodal fusion strategy that enhances accuracy; our approach yields a 92\% accuracy rate on a real-life trial dataset. Most important of all, the model indicates the attention focus in the videos, providing valuable insights on deception cues. Hence, our method adeptly detects deceit and elucidates the underlying process. We further enriched our study with an experiment involving students answering questions either truthfully or deceitfully, resulting in a new dataset of 309 video clips, named ATSFace. Using this, we also introduced a calibration method, which is inspired by Low-Rank Adaptation (LoRA), to refine individual-based deception detection accuracy.

\end{abstract}

\IEEEpeerreviewmaketitle

\section{Introduction}
Deception detection plays a crucial role in various domains, including court trials, job interviews, criminal investigations, and financial evaluations. Traditionally, trained experts would analyze an individual's micro-expressions, verbal characteristics, and transcriptions to determine the probability of deceitful behavior. However, recent advancements in artificial intelligence have led to the development of intelligent systems that are capable of acting as expert deception detectors. These systems have demonstrated remarkable accuracy rates, with some achieving up to 96.14\% on real-life trials dataset \cite{krishnamurthy2023deep}. Such advances have the potential to greatly enhance the efficiency and effectiveness of deception detection in various settings, leading to better outcomes for all involved parties.

Nonetheless, videos contain a vast amount of information in each frame or second, presenting a significant challenge due to the high-dimensional nature of the data and the distinct representations of visual and audio modalities. However, recent advancements in the field of multimodal fusion have led to the development of more sophisticated and accurate models for integrating information from diverse modalities. This progress has enabled the exploration of new applications in various domains, including robotics, human-computer interaction, and multimedia content analysis. As a result, multimodal fusion has become a crucial aspect of modern machine learning and has the potential to impact various fields by enabling more robust and accurate decision-making systems.

Aside from the aforementioned challenges, there are several other obstacles when analyzing video data. First, video data often come in varying lengths, which makes it difficult for models with fixed-size inputs to process the information effectively. Moreover, videos can contain a vast range of emotions, gestures, and expressions that are challenging to capture accurately. Additionally, variability in camera angles, lighting, and other environmental factors can cause significant differences in the video's quality, making it difficult to extract relevant information for deception detection. Furthermore, video data can come in various formats, resolutions, and compression levels, which may affect the quality and consistency of the data. Analyzing video data requires preprocessing and normalization steps to address these variations and ensure that the data is suitable for analysis.

However, a significant limitation of these AI-based deception detection models is that they work as a "black box". In other words, while the AI model can determine whether an individual is being deceptive or truthful, it often fails to provide a clear explanation or reasoning behind its judgment. This lack of interpretability poses a challenge for analysts to understand the factors that contribute to the AI model's decision-making process in deception detection. Consequently, there is a growing demand for more transparent and explainable AI models that can not only accurately detect deception but also offer insights into the underlying reasons for their assessments.

Deception can be viewed as a complex "process", suggesting that an interviewee is not necessarily lying throughout the entire conversation. This perspective highlights the dynamic nature of deception, where individuals may switch between truthfulness and dishonesty depending on the context, their intentions, and the information being discussed. Consequently, our model must continuously evaluate both the current information and past context to determine the final outcome. To address these challenges, we propose a recurrent neural network that incorporates an attention mechanism across multiple resources. 

In our study, we introduce an attention-aware neural network designed to identify the most crucial aspects of visual, audio, and transcription data for deception detection. We present an attention mechanism designed to continuously assess facial, voice, and textual information, identifying specific moments in video, audio, and text data that reveal signs of deception. Moreover, we embrace a multimodal approach by incorporating an ensemble mechanism following multiple models with distinct features, allowing for collaborative inference of the results. This strategy enables different models to offer varying perspectives for more accurate and comprehensive deception detection.

In addition, we introduced a calibration method, which is inspired by Low-Rank Adaptation (LoRA) \cite{hu2021lora}, to refine individual-based deception detection accuracy. The design rationale is that we notice different individuals may reveal different characteristics while lying, and it might be difficult map all individuals into a single latent space. Therefore, we introduce an extra neural network for each one to re-map the data into a single latent space.

Most important of all, we conducted an experiment with university students who were instructed to answer general questions about their school life and financial matters truthfully, as well as to create fictitious narratives on various personal topics. The resulting dataset, which we have named ATSFace, comprises 309 videos, evenly distributed between deceptive and truthful clips. This dataset was supplemented with detailed transcripts, generated through an automatic speech recognition system. Further details about the experiment and our approach to data collection and processing are presented in Section 4.

In our experiments, our proposed model demonstrates a remarkable 92.00\% accuracy rate when applied to a court trial dataset and 79.57\% accuracy rate on our own dataset. This performance is comparable to other research efforts that utilize the same dataset and feature extraction methods. As anticipated, our multimodal ensemble model yields superior results compared to unimodal approaches, emphasizing the benefits of fusing diverse sources of information in deception detection.

Furthermore, our model is designed to provide valuable insights for analysts by simultaneously outputting attention weights for each moment during the analysis. This feature enables analysts to identify specific time intervals that may contain critical cues related to deception. This combination of high accuracy and interpretability makes our model a powerful tool for both detecting deception and understanding its underlying dynamics in various contexts.
\section{Related Work}

Over the past few years, the field of deception detection has experienced rapid advancements. In 2015, a novel public dataset was introduced for deception detection, derived from real-life trial video data \cite{perez2015deception}. This dataset comprises 121 video clips, encompassing both verbal and non-verbal features. In their initial work with the dataset, the researchers focused on investigating the potential of non-verbal features (i.e., micro-expression) for deception detection. By employing machine learning algorithms such as Decision Tree (DT) and Random Forest (RF), they could classify deceptive behavior with an accuracy of 75\% with verbal and non-verbal features.

Subsequently, reference \cite{gogate2017deep} proposed a fully-connected layer-based model for automated deception detection. This model incorporated audio features using the openSMILE library, visual features through a 3D-CNN model, and text features via Text CNN. They implemented both early fusion and late fusion, discovering that the early fusion model performed better, achieving deception prediction accuracy of up to 96\%. Since this development, multimodal research has been extensively applied to court trial datasets.

In \cite{wu2018deception}, they utilized Improved Dense Trajectory (IDT) to extract visual features by computing local feature correspondences in sequential frames. They applied Mel-frequency Cepstral Coefficients (MFCCs) with Gaussian Mixture Model (GMM) for audio features, and Global Vectors for Word Representation (Glove) for transcription features. Additionally, they trained a linear kernel SVM to detect micro-expression as another feature for classification. After feature encoding, they tested several classification algorithms, including SVM, DT, RF, Adaboost, etc.

In \cite{krishnamurthy2023deep}, they employed 3D-CNN, openSMILE, and Text-CNN to process visual, audio, and textual features independently. Then, a simple fully-connected neural network was trained to reduce the dimension. In the fusion part, they tried different fusion methods to map the feature vectors into a joint space. After their experiments, they used the Hadamard product on feature vectors, concatenated the resulting vector with micro-expression labels, and finally input it into a hidden layer with ReLU activation for classification. Ultimately, their approach achieved an accuracy rate close to 96\%.

In contrast to previous works, reference \cite{karimi2018toward} initially employed CNNs followed by an LSTM network on both visual and audio feature extraction. Additionally, they implemented an attention mechanism on visual cues in each frame, highlighting deception-related cues. Then, they concatenated the feature vectors and applied a non-linear activation function. For deception classification, they utilized Large Margin Nearest Neighbor (LMNN) \cite{weinberger2009distance}, a metric learning approach in k-Nearest Neighbor (kNN) classification.

In \cite{ding2019face}, they addressed data scarcity in the dataset by leveraging meta-learning and adversarial learning techniques. They primarily focused on visual frame sequences during feature extraction and employed ResNet50 \cite{he2016residual} as the backbone model to process facial expressions and body motions represented by optical flow maps. They introduced a cross-stream fusion architecture for these two features in their paper. By combining these methods, they trained an end-to-end deception detection model, achieving an accuracy of about 96\% using only visual features. When audio and textual features were incorporated, the accuracy increased to 97\%.

More recently, reference \cite{mathur2020introducing} adopted the state-of-the-art AffWildNet model \cite{bhamare2020deep}, consisting of CNN and GRU layers, to extract facial affect features. Additionally, they used OpenFace, openSMILE, and Linguistic Inquiry and Word Count (LIWC) to extract visual, audio, and textual features, respectively. After feature extraction, they developed an SVM model for unimodal analysis and an Adaboost model for ensembling. In contrast to the aforementioned studies, \cite{avola2020lietome} utilized the OpenPose library to extract hand gesture features, offering a unique approach to deception detection.
\section{Method}
\begin{figure*}[htbp]
    \centerline{\includegraphics[trim = {0mm 0 0 0}, clip, width=0.96\textwidth]{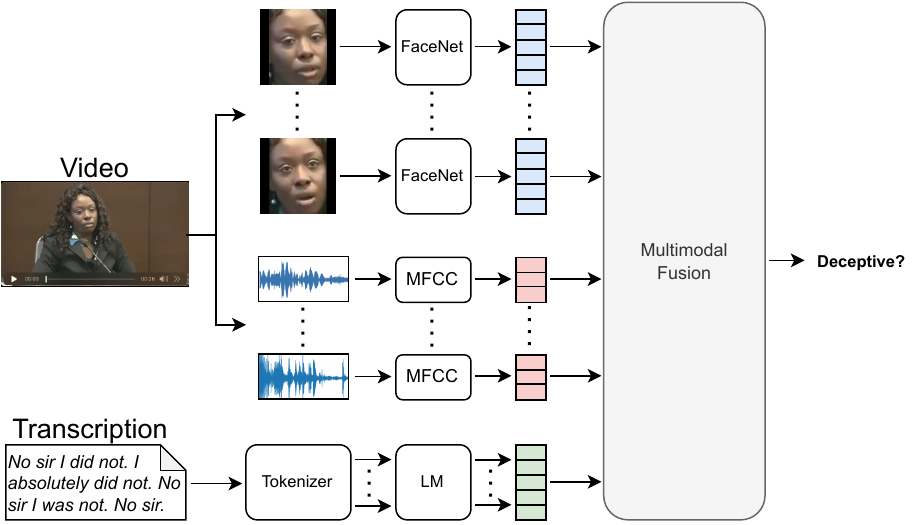}}
    \caption{Overview of our framework.}
    \label{fig:architecture}
\end{figure*}

Our model is composed of various components: (1) extracting multimodal features such as visual, audio, and transcription; (2) models based on these extracted features for the purpose of deception classification, as shown in Fig. \ref{fig:architecture}. (3) an attention mechanism to interpret visual, audio, and transcription features; and (4) an additional LoRA-like calibration network for improving individual deception detection accuracy . 

\subsection{Multimodal Feature Extraction}

\subsubsection{Visual Features} 
The extraction of visual features in our approach involves two main steps: detecting faces and converting them into vector representations. We employ RetinaFace \cite{deng2020retinaface} for face detection, which is a state-of-the-art facial detection algorithm that excels in detecting faces with high accuracy, even under challenging conditions such as small, blurry, or partially blocked faces. Subsequently, we utilize a pretrained FaceNet \cite{schroff2015facenet} model as the backbone, which leverages the Inception architecture, enabling it to learn a mapping of face images directly to a compact Euclidean space. As such, it enables us to transform detected faces into 128-dimensional vectors. Considering the variable lengths of videos and the differing number of frames, which complicates LSTM training, we opt to extract features at \emph{k}-frame intervals. This approach helps to generate a consistent representation of visual features while maintaining the essential information required for effective LSTM training.

\subsubsection{Audio Features} 
We employ MFCCs as our audio feature, which are a widely-used feature extraction technique in the field of speech and audio signal processing. They provide a compact representation of audio signals by capturing the spectral characteristics of the sound. MFCCs are derived from the cepstral analysis, which is a process that transforms the frequency domain of the audio signal into a time-like domain, focusing on the spectral shape rather than its amplitude. Similar to the challenge faced with visual features, training an LSTM directly on MFCCs with variable lengths can be difficult. To achieve this, we compute the mean of MFCCs per \emph{t} second, which not only reduces the length of the MFCCs representation but also enables the LSTM to be trained more effectively. This condensed representation maintains the essential information while allowing the LSTM to learn temporal patterns and relationships within the audio features more efficiently.

\subsubsection{Transcript Features} 
For the text transcription of each video, we employ a tokenizer as an encoding step, which is then passed through a language model. For English transcripts, we segment the text sentence by sentence, and each sentence is subsequently processed through the pretrained fastText \cite{bojanowski2017enriching}, generating a 100-dimensional vector per sentence. For Chinese transcripts, we employ the Chinese BERT tokenizer to process the text word by word. These vectors are then fed into Chinese BERT, pretrained by CKIP Lab, which generates a 768-dimensional vector for each word.

\subsection{Network Architecture}
After feature extraction, we implement multimodal fusion for the following prediction. In this section, we design two architectures: (1) late fusion (decision-level) and (2) multi-head cross-attention. Figure \ref{fig:architecture} provides an overview of our approach. Since the variable length of video data, we utilize Bidirectional LSTM (BiLSTM) as the primary model architecture. BiLSTM can effectively capture both past and future contexts of an input sequence, enabling the model to handle long-term dependencies better. Furthermore, because deception is a "process", we design an attention layer to capture the most informative features. The attention layer can help the model focus on specific moments of the video that reveal the deception cue. We will now provide a detailed description of each component.


\subsubsection{Unimodal}
The unimodal models comprise three main components: BiLSTM layers, an attention layer, and fully-connected layers. For each modality, we denote the extracted features as $x_i = \{x_i^m: 1 \leq i \leq N\}, m \in \{V, A, T\}$, where $m$ stands for visual, audio, or transcription feature. These features are initially processed through the BiLSTM layers, resulting in a sequence of hidden state vectors, $v_i$. The final hidden state vector, $v_i$, is formed by concatenating two vectors, which are computed by the forward and backward LSTM, as shown in the following equations.

\begin{equation}
    \begin{aligned}
        \overrightarrow{v_i} &= \overrightarrow{LSTM}(x_i) \\
        \overleftarrow{v_i} &= \overleftarrow{LSTM}(x_i) \\
        v_i &= [\overrightarrow{v_i}, \overleftarrow{v_i}] 
    \end{aligned}
    \label{bilstm}
\end{equation}

To gain insights into the deception detection mechanism of our model, we investigate the attention mechanism applied to visual, audio, and transcription features. Rather than focusing on the local feature details, we direct our attention to the frames of the videos. As aforementioned, we position the attention layer after the BiLSTM layer. For comparative analysis of the attention mechanism, we employ two distinct attention methods, namely, simple attention and scaled dot-product attention. During the attention calculation process, we compute attention scores and context vectors, providing a comprehensive understanding of the machine's decision-making process in deception detection. 

\paragraph{Simple Attention Layer}
The simple attention layer operates without the need for query, key, and value components typically utilized in an attention mechanism. In our model, we employ a weight matrix denoted by $W \in \mathbb{R}^{d_{model} \times 1}$ and a bias vector represented as $b \in \mathbb{R}^{N \times 1}$ to determine the relevance of each component in the input sequence.  The calculation process for the attention mechanism consists of the following equations:

First, we compute the hidden representation $h_i$ as follows:
\begin{equation}
    h_i = \tanh(v_{i} \cdot W) + b
    \label{tanh}
\end{equation}

Next, we determine the attention scores $\alpha_i$ using the softmax function:
\begin{equation}
    \alpha_i = \frac{exp(h_i)}{\sum_{i}{exp(h_i)}}
    \label{softmax}
\end{equation}

Finally, we calculate the context vector $c$ as follows:
\begin{equation}
    c = \sum_{i} v_{i}*\alpha_i
    \label{context}
\end{equation}

where $v_i$ stands for the output of the BiLSTM layers. The attention score for each frame $i$ is symbolized by $\alpha_i$, with $\alpha \in \mathbb{R}^{N \times 1}$. This 1-dimensional vector allows for the straightforward identification of the frames that the model focuses on. The context vector serves as the output of the attention layer and the input for the subsequent fully-connected layer. This process allows the model to weigh the importance of different moments in the input data when making decisions for deception detection. The output of the fully connected layers is passed through a 2-dimensional softmax function, which produces the final predictions.

\paragraph{Scaled Dot-Product Attention}
In contrast to the simple attention layer, the scaled dot-product attention incorporates query $Q$, key $K$, and value $V$ components. The function is defined as follows:

\begin{equation}
    Attention(Q, K, V) = softmax(\frac{QK^T)}{\sqrt{d_k}})V
\end{equation}

In this function, the query and the key are the BiLSTM output $v_i$. Then, the dot-product of the query and the key is scaled by the square root of the dimension of the key $d_k$. This value is then passed through the softmax function to obtain the weights on the value elements. By performing this operation, the model generates a weighted sum and the attention score, which are $N \times d_{model}$ and $N \times N$ dimensions respectively. As with the simple attention layer, to feed the output of the attention layer into the subsequent fully-connected layers, we employ a global max pooling operation, ultimately leading to the final deception detection prediction after the softmax operation.

By leveraging these two attention mechanisms, we can effectively account for the varying importance of different moments in the input data. Consequently, our model is able to make well-informed decisions for deception detection, facilitating a robust understanding of its decision-making process.

\subsubsection{Late Fusion}
The late fusion approach aims to create an integrated system that leverages the unique strengths of unimodal models for visual, audio, and transcription features. We initially build these models separately. Following this, an ensemble mechanism can effectively combine the insights obtained from the three models to yield the final result. We utilize the voting mechanism to achieve this fusion.

The voting mechanism represents the straightforward way of ensemble techniques. Each unimodal model independently evaluates the input data and generates its prediction. These individual predictions are then aggregated, and the final decision is made based on the majority rule, i.e., the outcome that receives the most votes from the three models is selected. This mechanism requires no additional training, as it merely collects and analyses the results of the existing models. It operates under the assumption that the majority of models will make the correct decision.

\subsubsection{Cross-Attention Fusion}
\begin{figure}[htbp]
    \centerline{\includegraphics[trim = {0mm 0 0 0}, clip, width=0.44\textwidth]{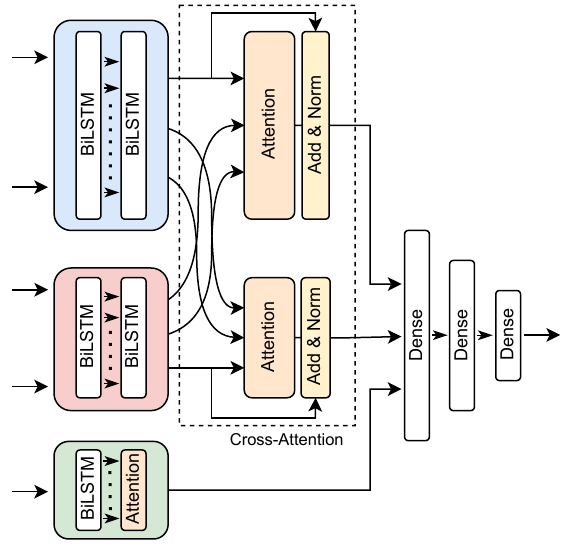}}
    \caption{Cross-Attention Architecture.}
    \label{fig:crossatt}
\end{figure}

To further explore the relationship between visual and audio features, we employ a cross-attention mechanism facilitated by scaled dot-product attention. As illustrated in Fig. \ref{fig:crossatt}, visual and audio features are initially processed through distinct BiLSTM layers as described in Equation \ref{bilstm}, which produce hidden state vectors $v^V$ and $v^A$. In the cross-attention process, we use $v^V$ as the query and $v^A$ as the key and value, inputting them into the scaled dot-product attention of the visual part, denoted as ${CA}_V$. Conversely, for the audio component, we generate ${CA}_A$ using the set {$v^A, v^V, v^V$}. Subsequently, we apply a residual connection \cite{he2016residual} and layer normalization \cite{hochreiter2001gradient}. The transcription model mirrors the process mentioned above. Finally, we feed these vectors into fully-connected layers to generate the ultimate prediction.

\subsection{LoRA-like Calibration}
\begin{figure}[htbp]
    \centerline{\includegraphics[trim = {0mm 0 0 0}, clip, width=0.45\textwidth]{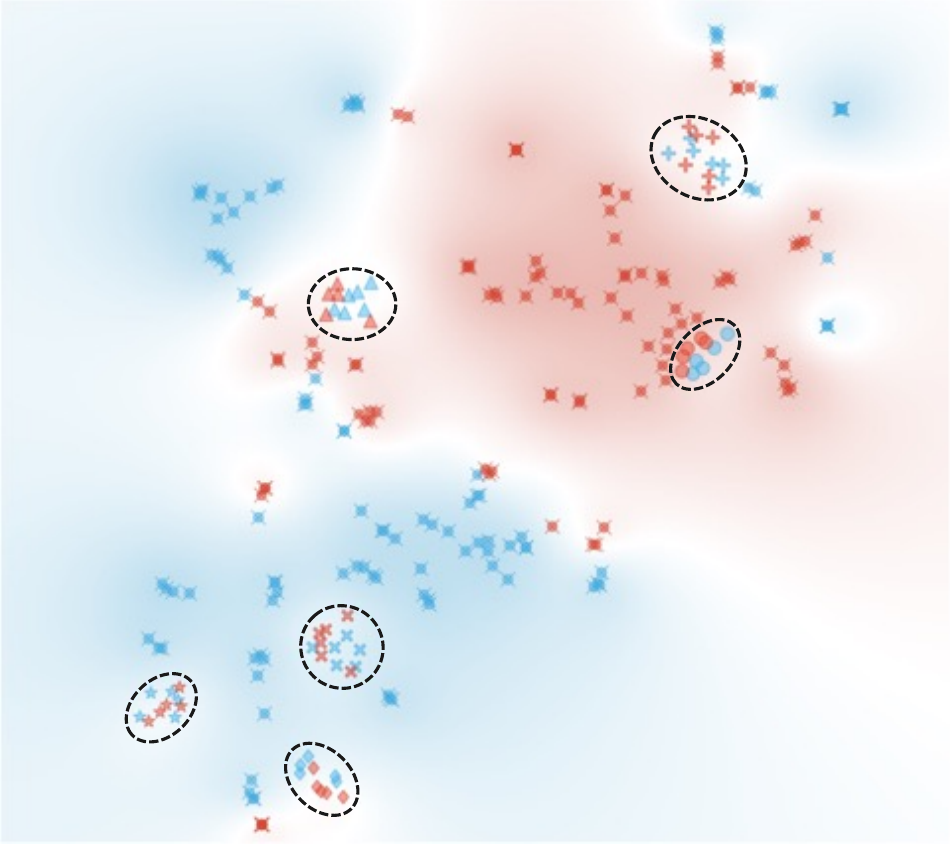}}
    \caption{Latent Space. Blue: Truthful; Red: Deceptive; Green circle: New people}
    \label{fig:latent}
\end{figure}

When a person tells a lie, the facial expression, voice, and words expressed can differ significantly based on the individual's characteristics. In previous experiments, we aimed for a generalization model, training it based on the lying behaviors common to most individuals. However, to understand the uniqueness of each person's deception cues, we take a subset of individuals from our dataset, train our model using the clips of the remaining individuals, and then test the model on the subset. We extract the output of the visual model's attention layer for all clips and plot these as latent vectors, as demonstrated in Fig. \ref{fig:latent}. Here, we observe that the extracted vectors tend to cluster together by the individual, but these clusters cannot be accurately classified by the base model. Therefore, we introduce LoRA structure to enhance the model's performance in individual deception detection.

\begin{figure}[htbp]
    \centerline{\includegraphics[trim = {0mm 0 0 0}, clip, width=0.4\textwidth]{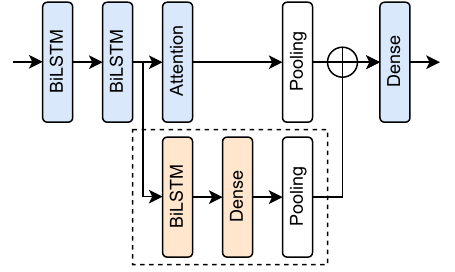}}
    \caption{LoRA-like Arichitecture}
    \label{fig:lora}
\end{figure}

Reference \cite{hu2021lora} presented the Low-Rank Adaptation (LoRA) method. This approach freezes pre-trained model weights and incorporates trainable matrices into each layer of the Transformer architecture. During training, the LoRA method indirectly trains specific dense layers in the neural network by optimizing the rank decomposition matrices, reflecting changes in these dense layers during adaptation. Consequently, LoRA enhances training efficiency and minimizes the number of required parameters. In order to adapt the model to these new people in our dataset, we design a LoRA-like model, as shown in Fig. \ref{fig:lora}.

As mentioned, we construct a new model composed of two BiLSTM layers, a scaled dot-product attention layer, a pooling layer, and multiple dense layers. These layers' weights are then frozen as pretrained weights, and we add a new BiLSTM layer, a time-distributed dense layer, and a pooling layer after the second BiLSTM layer of the pretrained model. Ultimately, we combine the outputs of the two pooling layers and pass them through the dense layers for classification. Unlike fine-tuning the base model, the introduction of the LoRA technique can prevent the original training data from being influenced by the new data since the weights of the base model remain unchanged. The LoRA-like model linearly transforms only the new data, folding them at points in the latent space so that the pretrained dense layers can achieve accurate classification.

\section{Evaluation}


\subsection{Dataset}
In deception detection research, the most commonly used existing dataset is the real-life trial video data in \cite{perez2015deception}. However, this dataset presents certain limitations, particularly its limited size and inconsistent recording quality. In our study, we create a new dataset, called ATSFace\footnote{GitHub link: \url{https://github.com/dclay0324/ATSFace}}, by experimenting with a multimodal approach to deception detection, which is described as follows. 

\subsubsection{Data Collection}
\begin{figure}[htbp]
    \centerline{\includegraphics[trim = {0mm 0 0 0}, clip, width=0.48\textwidth]{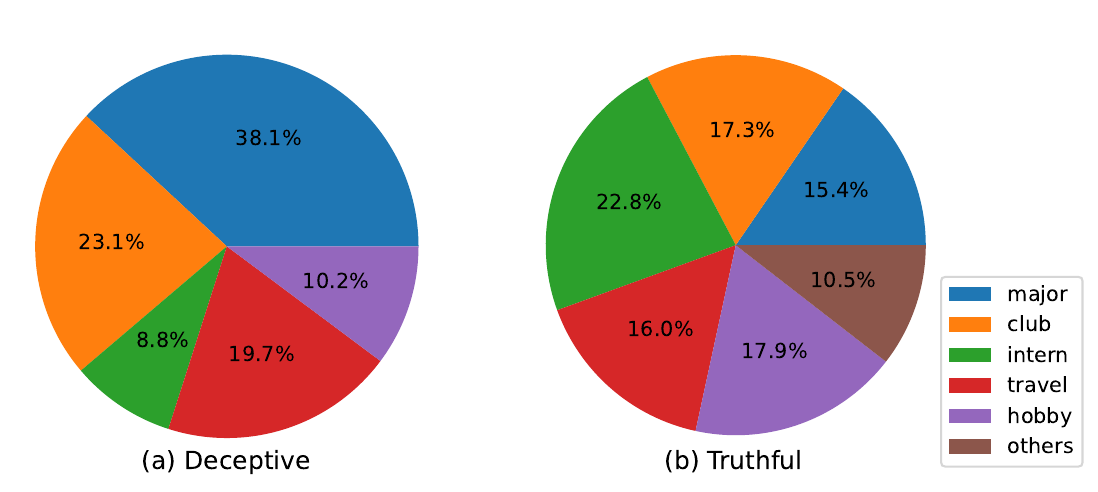}}
    \caption{Proportion of Subjects}
    \label{fig:subjects}
\end{figure}

The primary participants for this experiment are university students. Initially, they are posed with general questions about school life and finance, to which they are instructed to respond truthfully. Subsequently, they are asked to select a topic from a list that includes their major, club experiences, internships, travel experiences, and personal hobbies. Participants are instructed to create a fictitious narrative about the selected topic, detailing events or experiences they have never actually experienced. Finally, the participants are asked to choose another topic and provide an honest narrative. As shown in Fig. \ref{fig:subjects}, we present the proportion of subjects chosen by the participants.

The following is the example questions of major topic:
\begin{enumerate}
    \item What major are you currently studying? What are the primary courses you are taking in this major?
    \item Among these courses, which one is your favorite, and why?
    \item Can you describe the content and key learning points of your favorite course?
    \item Can you describe the teaching style of the professor of this course and his or her grading criteria?
    \item What would you tell an incoming freshman who asked you for advice on studying in the major?
\end{enumerate}

\begin{figure*}[htbp]
    \centerline{\includegraphics[trim = {0mm 0 0 0}, clip, width=0.96\textwidth]{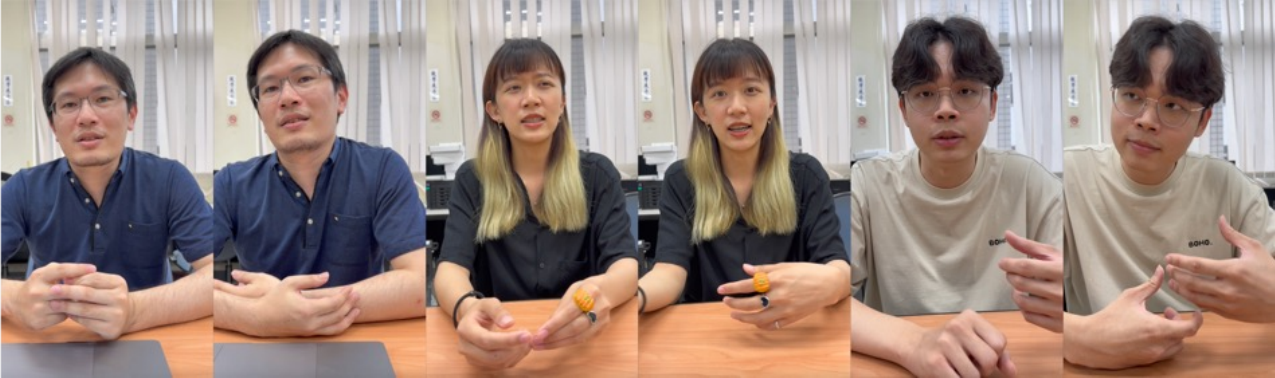}}
    \caption{Sample Screenshots of Videos}
    \label{fig:video}
\end{figure*}

For the experiment, we employed an iPhone 14 Pro for recording in a 1080p HD/30fps format, with the device positioned upright. Participants were instructed to stay seated and respond to the moderator's questions in Chinese for the entire experiment duration. Figure \ref{fig:video} exhibits screenshots showcasing various facial expressions captured from the video clips, demonstrating behaviors such as head movements, scowling, and upward eye gazes, among others.

\begin{figure}[htbp]
    \centerline{\includegraphics[trim = {0mm 0 0 0}, clip, width=0.48\textwidth]{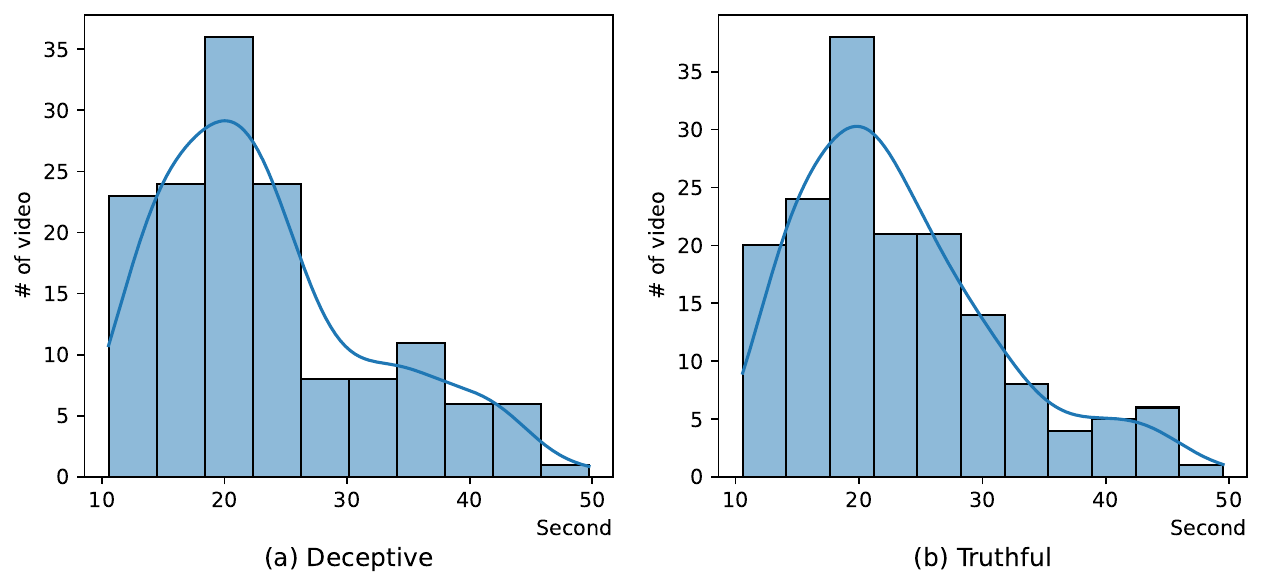}}
    \caption{Distribution of Video Length}
    \label{fig:distribution}
\end{figure}

The final dataset derived from the experiment consists of 309 videos, of which 147 are deceptive and 162 are truthful clips. The average duration of these videos is 23.32 seconds, ranging from 10.53 to 49.73 seconds. The average lengths for deceptive and truthful clips are 23.33 seconds and 23.30 seconds, respectively. The distributions of the video length are shown in Fig. \ref{fig:distribution}. We try to make the distribution of the two labels as exact as possible so that the model is not affected by the length of the videos during training and testing. The data consists of 23 unique male and 13 unique female speakers.

Table \ref{transcript} presents transcripts of sample deceptive and truthful statements from the dataset. We employ CapCut, an automatic speech recognition (ASR) system for transcription. We retain the filler and repeated words to preserve the originality of the text. The final collection of transcriptions comprises 35,069 words, including 1,403 unique words, averaging 113 words per transcript.

\begin{table*}[htbp]
    \caption{Sample Transcripts}
    \begin{center}
        \begin{tabular}{|c|p{7.5cm}|p{7.5cm}|}
            \hline
            & \textbf{Deceptive} & \textbf{Truthful} \\
            \hline
            \multirow{6}{*}{English} & 
            The teacher, Dr. Yang, is, um, a doctor at the hospital. He primarily teaches how to perform cardiac surgeries. Although, um, his grading tends to be lenient, the course is, um, an anatomical and medical course, so he is quite strict. This is so we won't encounter some, um, problems during surgery in the future. & 
            In terms of accomplishments, at various stages, for instance, um, during high school, I participated in clubs, or there were some science fairs during high school. So, I might have made some music for these events, or um, for these clubs. Or if they had some music-related activities, I would assist them, either as a performer or as someone providing the background music. \\
            \hline
            \multirow{5}{*}{Chinese} & 
            \begin{CJK}{UTF8}{bsmi} 
            這位老師姓楊是呃醫院的楊醫生，他主要教的內容就是如何做心臟的外科手術，那呃老師分數偏甜，但是因為是解剖的呃是醫學的課程，所以呃還是蠻嚴厲的，因為這樣我們以後才不會出現一些呃手術上的問題。
            \end{CJK} & 
            \begin{CJK}{UTF8}{bsmi}
            目前成就的話，就是我會在各個階段的時候，比如說呃在高中時期有參加社團，或者是高中時期有一些科展，那可能就是替這個活動或者說替這個社團做一些音樂，或者是呃如果他們有一些音樂相關活動的話，我就是都會協助他們，就是當那個表演者啊，或是當就是配樂這樣子。
            \end{CJK} \\
            \hline
        \end{tabular}
        \label{transcript}
    \end{center}
\end{table*}

\subsubsection{Real-Life Trial}
We also use the real-life trial video data in \cite{perez2015deception} for evaluating our models. This dataset consists of 121 video clips, including 61 deceptive and 60 truthful trial clips. In this dataset, the videos have several fps from 10 to 30. Then, the videos are variable in length, ranging from 4.5 to 81.5 seconds, and the average length is 28 seconds. We split the dataset into two subsets, training (80\%) and testing (20\%), and we perform 10-fold cross-validation.

\subsection{Implementation Details}
\subsubsection{Max Padding Length}
The intervals for visual feature extraction are set to capture 5 frames per second, ensuring a consistent temporal representation across different videos, irrespective of their original frame rates. For instance, if the frame rate is 30, we select every sixth frame. In the case of audio feature extraction, we compute the mean MFCCs every $t=0.2$ second to align with the visual features' interval.

In the process of visual feature extraction, there are instances where we cannot detect a clear face in certain frames in the real-life trials dataset. To address this issue, we pad the frame with the previously detected face and continue to do so until the next face is detected.  After extraction, we pad all three modalities' features to the maximum feature length.

\subsubsection{Model Training and Parameters}
In the real-life trials dataset, we set the BiLSTM layer units to 64 and 32 for both visual and audio models, and to 32 for the textual model. For the dense layers, we use 64, 16, and 8 units for the visual and textual models, 32, 16, and 8 units for the audio model, and 256, 64, and 16 units for the stacking and cross-attention model. The number of epochs is set to 20 for the visual model, 30 for the audio and textual models, and 15 for the stacking and cross-attention models.

In our ATSFace dataset, the settings differ slightly: the BiLSTM layer units are set to 64 and 32 for the visual and audio models, and to 128 and 64 for the textual model. The dense layer units for all three unimodal models are uniformly set to 64, 16, and 8, and for the stacking and cross-attention model, they are set to 256, 64, and 16. The number of epochs is 40 for the visual and audio models, 30 for the textual model, and 20 for the stacking and cross-attention model. For all models, the batch size is 32. Finally, we employ the Adam algorithm as our optimizer.

\subsubsection{Learning Rate Scheduler}
We apply a learning rate scheduler to adjust the learning rate during training. The learning rate remains constant for the first \emph{N} epochs. After the \emph{N} epoch, the learning rate decreases exponentially with a decay factor of 0.1 for each subsequent epoch. This decay scheme is formally expressed by the following equation:

\begin{equation}
    lr_{new} =
    \begin{cases}
        lr_{old}, & \text{if } n < N \\
        lr_{old} \times e^{-0.1}, & \text{otherwise}
    \end{cases}
    \label{lrscheduler}
\end{equation}

where $lr_{new}$ is the updated learning rate, $lr_{old}$ is the previous learning rate, and $n$ denotes the current training epoch. The learning rate is multiplied by $e^{-0.1}$ after the \emph{N} epoch to gradually reduce the step size, promoting model convergence in the optimization landscape.

In the real-life trials dataset, we set the number of epochs $N$ to 10 for the visual model, 20 for the audio model, 25 for the text model, and 10 for both the stacking and cross-attention model. In our ATSFace dataset, we adjust $N$ to 20 for the visual and textual model, 30 for the audio model, and 10 for both the stacking and cross-attention model.

\subsection{Experiment}

\subsubsection{Results on Real-life Trials Dataset}
The experiment results are presented in Table \ref{tab1}. In the unimodal section, it can be seen that the visual model implementing two attention methods achieves the highest accuracy at 88.80\%, although the scaled dot-product attention method boasts a higher F1-score. For the audio and textual models, both perform well with simple attention. In the multimodal fusion section, the voting mechanism reaches the highest accuracy at 92.00\% with an F1-score of 91.90\%. 

\begin{table}[htbp]
    \caption{Prediction Accuracy on Real-life Trials Dataset}
    \begin{center}
        \begin{tabular}{|c|c|c|c|}
            \hline
            \textbf{Feature} & \textbf{Attention} & \textbf{Accuracy (\%)} & \textbf{F1-Score (\%)} \\
            \hline
            \hline
            \multirow{2}{*}{Visual} & Simple & 88.80 & 88.48 \\
            \cline{2-4}
            & Dot-Product & 88.80 & 88.73 \\
            \hline
            \multirow{2}{*}{Audio} & Simple & 84.80 & 84.46 \\
            \cline{2-4}
            & Dot-Product & 83.20 & 82.99 \\
            \hline
            \multirow{2}{*}{Text} & Simple & 77.60 & 77.14 \\
            \cline{2-4}
            & Dot-Product & 77.00 & 76.27 \\
            \hline
            \hline
            Voting & - & \textbf{92.00} & \textbf{91.90} \\
            \hline
            Cross-Attention & Dot-Product & 90.40 & 90.25 \\
            \hline
        \end{tabular}
        \label{tab1}
    \end{center}
\end{table}

Table \ref{tab2} showcases a comparative analysis between our models and the approach proposed in the referenced research, specifically focusing on studies that employ the same feature extraction methods as ours on this real-life trials dataset. Notably, we draw a comparison with the study by Karimi et al. \cite{karimi2018toward}, which like our research, implements an LSTM model. This comparison serves as a benchmark, affirming the validity and competitiveness of our model relative to the existing state-of-the-art approaches in deception detection.

\begin{table}[htbp]
    \caption{Accuracy Comparisons on Real-life Trials Dataset}
    \begin{center}
        \begin{tabular}{|c|c|c|c|c|c|}
            \hline
            \multirow{2}{*}{\textbf{Method}} & \multirow{2}{*}{\textbf{Metrics}} & \multicolumn{4}{|c|}{\textbf{Modality (\%)}} \\
            \cline{3-6}
            & & Visual & Audio & Text & Multimodal \\
            \hline
            \hline
            \cite{karimi2018toward} & ACC & 75.00 & 74.16 & - & 84.16 \\
            \hline
            \cite{mathur2020introducing} & ACC & 76.00 & 72.00 & 63.00 & 84.00 \\
            \hline
            \cite{wu2018deception} & AUC & 77.31 & 76.94 & 64.57 & 92.21* \\
            \hline
            \cite{gogate2017deep} & ACC & 78.57 & 87.5 & 83.78 & 96.42 \\
            \hline
            \hline
            Ours & ACC & 88.80 & 84.80 & 77.60 & 92.00 \\
            \hline
            \multicolumn{6}{l}{*Use micro-expression as feature.}
        \end{tabular}
        \label{tab2}
    \end{center}
\end{table}

\subsubsection{Results on ATSFace Dataset}
The experiment results on our own dataset are shown in Table \ref{tab3}. It is evident that our model demonstrates effectively in classifying deceptive and truthful clips when utilizing visual and textual features. These results indicate the substantial potential of our model to discriminate between truthful and deceptive instances based on visual cues and text content. However, the performance of our model decreases significantly when applied to the audio features of our dataset, indicating lower effectiveness in distinguishing deception based on audio cues.

\begin{table}[htbp]
    \caption{Prediction accuracy on ATSFace Dataset}
    \begin{center}
        \begin{tabular}{|c|c|c|c|}
            \hline
            \textbf{Feature} & \textbf{Attention} & \textbf{Accuracy (\%)} & \textbf{F1-Score (\%)} \\
            \hline
            \hline
            \multirow{2}{*}{Visual} & Simple & 73.12 & 72.70 \\
            \cline{2-4}
            & Dot-Product & 73.12 & 73.04 \\
            \hline
            \multirow{1}{*}{Audio} & Simple & 61.29 & 60.46 \\
            \hline
            \multirow{2}{*}{Text} & Simple & 76.88 & 76.69 \\
            \cline{2-4}
            & Dot-Product & 76.88 & 76.86 \\
            \hline
            \hline
            Voting & - & \textbf{79.57} & \textbf{79.23} \\
            \hline
            Cross-Attention & Dot-Product & 73.66 & 73.47 \\
            \hline
        \end{tabular}
        \label{tab3}
    \end{center}
\end{table}

\subsection{Visual Interpretability}
\begin{figure}[htbp]
    \centerline{\includegraphics[trim = {0mm 0 0 0}, clip, width=0.48\textwidth]{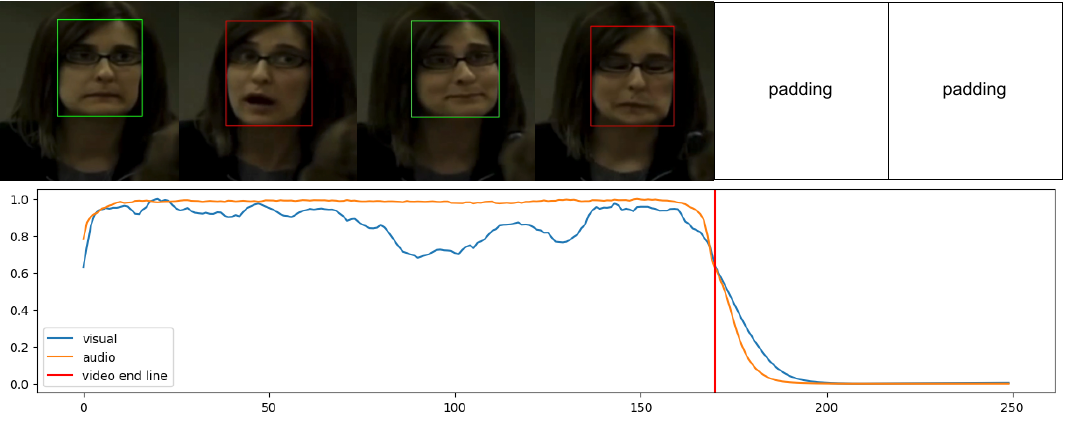}}
    \caption{Attention score on video}
    \label{fig:att}
\end{figure}

In order to display the attention results, we extract the variable $\alpha_i$, representing the attention score within the video. This variable allows us to understand the moments the model considers pivotal for deception detection. As depicted in Fig. \ref{fig:att}, the video is 34 seconds long, labeled as 'deceptive'. We sample one frame for every 6 frames, resulting in a rate of 5 frames per second, and yielding a total of 169 frames. Our goal is to identify which frames the AI model deems most critical in determining deception. Frames with the top $k$ attention scores are highlighted with a red rectangle around the face. Notably, during these moments, the interviewee shakes her head and compresses her lips, suggesting the AI model views these expressions as deceptive cues. Given that the video comprises 169 frames, attention scores drop sharply after frame 170, indicating the model recognizes the insignificance of padding values.

\subsection{LoRA-like Calibration}

\begin{figure*}[htbp]
    \centerline{\includegraphics[trim = {0mm 0 0 0}, clip, width=0.96\textwidth]{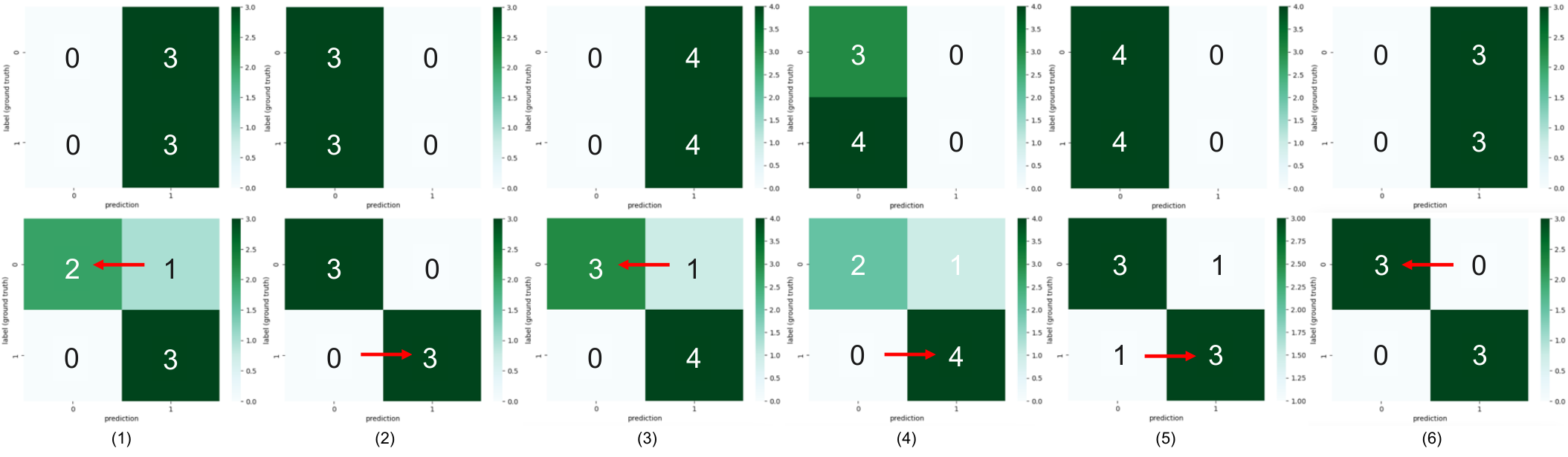}}
    \caption{Confusion Matrix. Upper: Base Model; Lower: LoRA-like model}
    \label{fig:cm}
\end{figure*}

In our experiments, we select a subset of 6 individuals, each with more than 5 lying and truthful clips, amounting to 65 clips in total, denoted as $S_\mathcal{L}$. We train the base model with the remaining 244 clips, which we divide into a training set (80\%) and a testing set (20\%), referred to as $S_\mathcal{R}$ and $S_\mathcal{T}$. After training, we plot the latent vectors of $S_\mathcal{L}$ and $S_\mathcal{R}$, as shown in Fig. \ref{fig:latent}. For $S_\mathcal{L}$, we calibrate the LoRA-like model by individuals, using 2 deceptive clips and 2 truthful clips for model training, with the remaining clips serving as a testing set. The accuracy on the testing clips for all 6 individuals was 48.78\%. We employ an early stopping mechanism during the training of the LoRA-like model, halting the process when training accuracy reached 100.0\%. After training, the accuracy on the testing clips for all 6 individuals increases to 87.80\%. Figure \ref{fig:cm} displays the confusion matrices for each individual, as derived from the two models. The effectiveness of the linear transformation is evident.

\section{Conclusion}

\begin{itemize}
    \item We propose an attention-aware recurrent neural network model architecture for deception detection, which effectively processes time-sequence video data, including visual, audio, and textual modalities. This model achieves remarkable accuracy in unimodal tasks.
    \item Our model adopts a multimodal fusion mechanism, enhancing detection accuracy and comprehensiveness by integrating diverse modalities. The multimodal approach outperforms unimodal methods, underscoring the advantages of incorporating multiple information sources.
    \item Our work introduces a new dataset, comprising 309 videos of university students' truthful and deceptive responses to various topics, supplemented with detailed automatic speech recognition transcripts. This dataset, with its clear recording of facial expressions and sounds, facilitates more accurate analysis.
    \item We design a LoRA-like model to calibrate our base model to individual characteristics. This strategy effectively improved our model's performance on a subset of individuals. The model preserved the base model's integrity, demonstrating its effectiveness in individualized deception detection.
\end{itemize}

\nocite{*}
\bibliographystyle{IEEEtran}
\bibliography{thesis}

\vspace{12pt}
\end{document}